\documentclass[10pt,twocolumn,letterpaper]{article}

\usepackage{cvpr}
\usepackage{times}
\usepackage{epsfig}
\usepackage{graphicx}
\usepackage{amsmath}
\usepackage{amssymb}
\usepackage{fancyhdr}

\usepackage[pagebackref=true,breaklinks=true,letterpaper=true,colorlinks,bookmarks=false]{hyperref}

\cvprfinalcopy % *** Uncomment this line for the final submission

\def\cvprPaperID{****} % *** Enter the CVPR Paper ID here

\ifcvprfinal\fi
\begin{document}

%%%%%%%%% TITLE
\title{Analysis \& Computational Complexity Reduction of \\ Monocular and Stereo Depth Estimation Techniques }

\author{Rajeev Patwari \thanks{The authors contributed equally}\\
%rpatwari@stanford.edu\\
{\tt\small rpatwari@stanford.edu}
\and
Varo Ly \footnotemark[1] \\
{\tt\small varoly@stanford.edu}
}

\maketitle
%\thispagestyle{empty}

%\footnote{The authors contributed equally}

%%%%%%%%% ABSTRACT
\begin{abstract}
Accurate depth estimation with lowest compute and energy cost is a crucial requirement for unmanned and battery operated autonomous systems. Robotic applications require real time depth estimation for navigation and decision making under rapidly changing 3D surroundings. A high accuracy algorithm may provide the best depth estimation but may consume tremendous compute and energy resources. A general trade-off is to choose less accurate methods for initial depth estimate and a more accurate yet compute intensive  method when needed. Previous work has shown this trade-off can be improved by developing a state-of-the-art method \cite{anynet} to improve stereo depth estimation. 

We studied both the monocular and stereo vision depth estimation methods and investigated methods to reduce computational complexity of these methods. This was our baseline. Consequently, our experiments show reduction of monocular depth estimation model size by $\sim$75\% reduces accuracy by less than 2\% (SSIM metric). Our experiments with the novel stereo vision method \cite{anynet} show that accuracy of depth estimation does not degrade more than 3\% (three pixel error metric) in spite of reduction in model size by ~20\%. We have shown that smaller models can indeed perform competitively. The source code is available at https://github.com/rajeevpatwari/AnyNet and https://github.com/rajeevpatwari/Monocular-Depth
\end{abstract}

%%%%%%%%% BODY TEXT
\section{Introduction}
The current state of the art autonomous driving and real time navigation decision making tasks require visual scene understanding as an addendum to radio or other sensing techniques.  Depth estimation in computer vision and robotics is a necessary and crucial step.  This is done today using stereo vision depth techniques \cite{stereo_survey}, monocular depth techniques \cite{wofk2019fastdepth}, and hybrid approaches \cite{saxena}.

%------------------------------------------------------------------------
\section{Background and Related Work}
%-------------------------------------------------------------------------
In our project, we implemented monocular and stereo depth estimation using supervised machine learning techniques.  Our motivation for the models chosen were to choose models that could provide adequate accuracy while also allowing less computation which would enable depth estimation on lower latency and lower power systems in real world applications.

\subsection{Monocular Depth}
Monocular depth estimation from a single image can be accomplished using classical supervised learning autoencoder techniques as in \cite{basuviktor}, \cite{wofk2019fastdepth}, and \cite{thatchervolk}.  The network in \cite{wofk2019fastdepth} was developed to provide both accuracy and low latency. Improvements to the speed and computational complexity of the these models were accomplished by utlizing an efficient pretrained encoder such as MobileNet and pruning the network using NetAdapt.  In our studies we utlized a basic CNN for both the enoder and decoder similar to \cite{basuviktor}.

\subsection{Stereo Depth}
In stereo depth estimation using supervised machine learning techniques, images from two cameras and ground truth disparity data are used to train a neural network to estimate the disparity.  In \cite{stereo_survey} many state of the art methods were reviewed.  The process of stereo depth estimation using these techniques generally consisted of feature extraction from both images, feature matching between the images, disparity estimation, and final refinement of the disparity map.

%-------------------------------------------------------------------------
%------------------------------------------------------------------------
\section{Technical Approach}

\subsection{Monocular Depth}

\subsubsection{Monocular Depth Architecture}
\begin{figure}[h]
	\includegraphics[width=\linewidth]{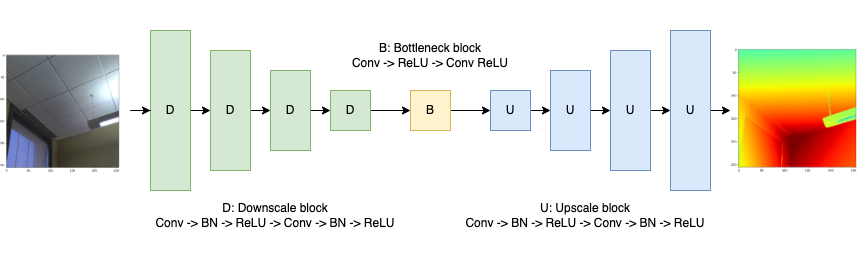}
	\caption{Monocular Depth Estimation Model}
	\label{fig:mon_nn}
\end{figure}

For monocular depth estimation, we reproduced a model based on \cite{basuviktor}.  This consisted of creating a neural network as shown in Fig \ref{fig:mon_nn} consisting of four downscaling blocks, a bottleneck block, and 4 upscaling blocks.

\subsubsection{Monocular Evaluation Metric}

We used three loss functions in our monocular depth estimation technique.  The following equation shows L1 loss used for model optimization.
$L1_{loss} = \sum_{i=1}^{n} |d_{i,true}-d_{i,pred}|$ 
In addition to L1 loss, we also used depth smoothness and SSIM \cite{ssim} per Tensorflow.  The model training involved minimizing all three losses.

We evaluated the accuracy of our generated depth maps by comparing them with the target depth maps using the structural similarity index metric.

\subsection{Stereo}

\subsubsection{Stereo Model Architecture}

\begin{figure}[h]
	\includegraphics[width=\linewidth]{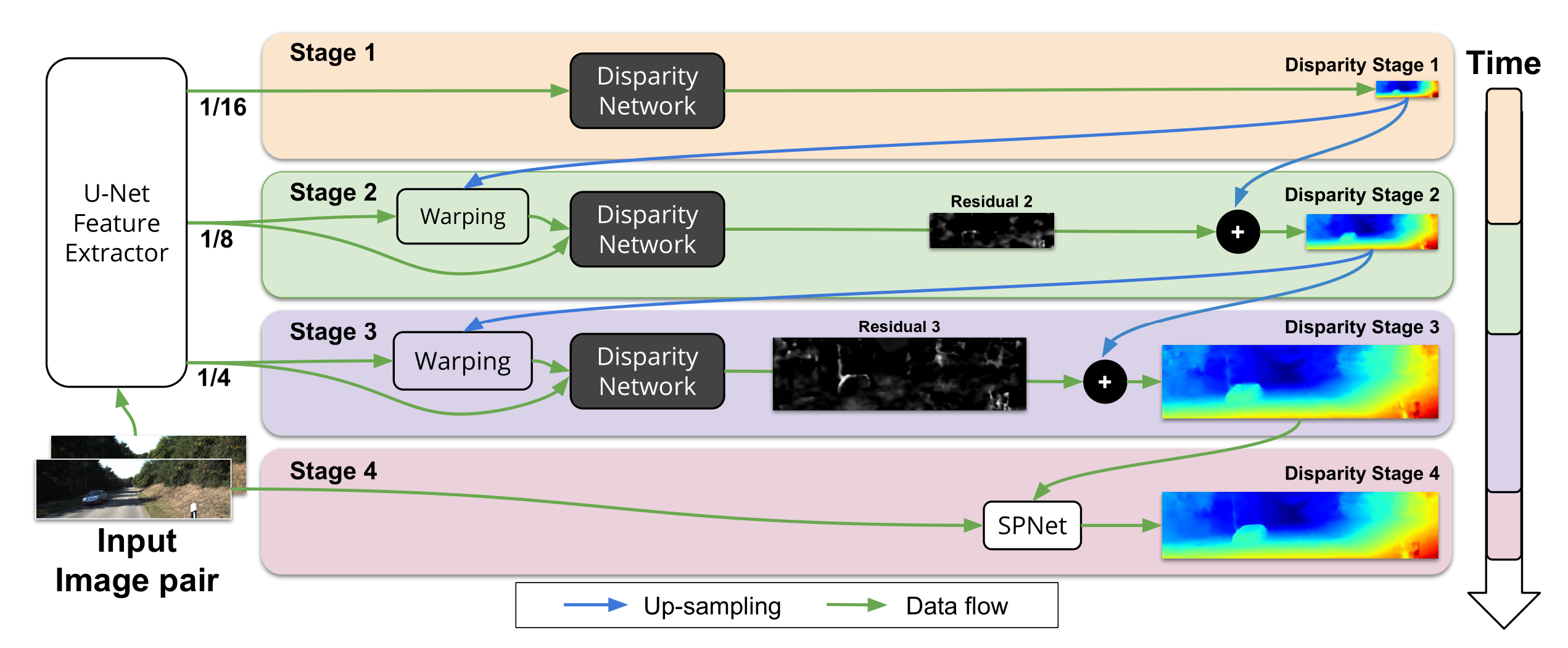}
	\caption{AnyNet \cite{anynet} Model}
	\label{fig:anynet_model}
\end{figure}

Based on our review of \cite{stereo_survey}, we chose to reproduce a stereo vision technique using the AnyNet \cite{anynet} model.  We chose this model as it was fast, accurate, and one of the computationally less intensive models.  Figure \ref{fig:anynet_model} depicts the architecture of the model which consists of the typical components as previously discussed, except they are combined over 4 different stages.

The U-Net feature extractor is an approximately 20 layer CNN that extracts the features from both the left and right images at various scales (1/16, 1/8, and 1/4) of the images.  The warping block matches the features between the left and right images.  The disparity network CNN for each stage is approximately 7 layers.  For the first stage since the image scale is so small, there is no feature matching before the disparity network CNN.    The refinement component is accomplished by an additional CNN called the spatial propagation network, SPNet, which is applied to the predicted disparity map following stage 3 to further improve the results.  As part of our studies, we varied the size and complexity of SPNet and assessed it's impact on the accuracy of the results.

\subsubsection{Stereo Evaluation Metric}
The smooth L1 loss was during training for stereo depth estimation.  The smooth L1 loss corresponds to the L2 loss when the difference between the outputs is below a specified parameter and corresponds to the L1 loss when it is above this chosen value.  This can be tuned as a hyper parameter, $\beta$.
\[ L1_{smooth\_loss} =  \]
\[
\sum_{i=1}^{n} \begin{cases}
\frac{1}{2\beta}(d_{i,true}-d_{i,pred})^2 & \text{if\;} |d_{i,true}-d_{i,pred}| < \beta \\
|d_{i,true}-d_{i,pred}|-.05\beta & \text{otherwise}
\end{cases}
\]

To assess our stereo model results, we utilized a three pixel error criteria as described in \cite{anynet} and supported code {\tt\small \url{https://gist.github.com/MiaoDX/8d5f49c2ccb39d7f2cb8d4e57c3ab752}}. 

%-------------------------------------------------------------------------
%------------------------------------------------------------------------
\section{Experiments and Results}
\subsection{Dataset}
\subsubsection{Monocular Dataset}
KITTI \cite{Geiger2013IJRR}, NYU-V2 \cite{Silberman:ECCV12}, DIODE \cite{diode_dataset} are three datasets available for monocular depth estimation, but all three datasets are approximately 100 GB.  To reduce training time, we have used a validation set from the DIODE dataset which comprises of 325 indoor depth maps and 446 outdoor depth maps.  And each data sample comprises of RGB image, depth map, and depth validity mask.  Each image is 1024x768, however to reduce the training time we reduced the image size to 256x256 and consequently clipped the depth map as well that is provided as a label to the network.  Using this technique each epoch takes approximately 10 minutes to train on the GTX1070 and V100.

\begin{figure}[h]
	\centering
	\includegraphics[width=2in]{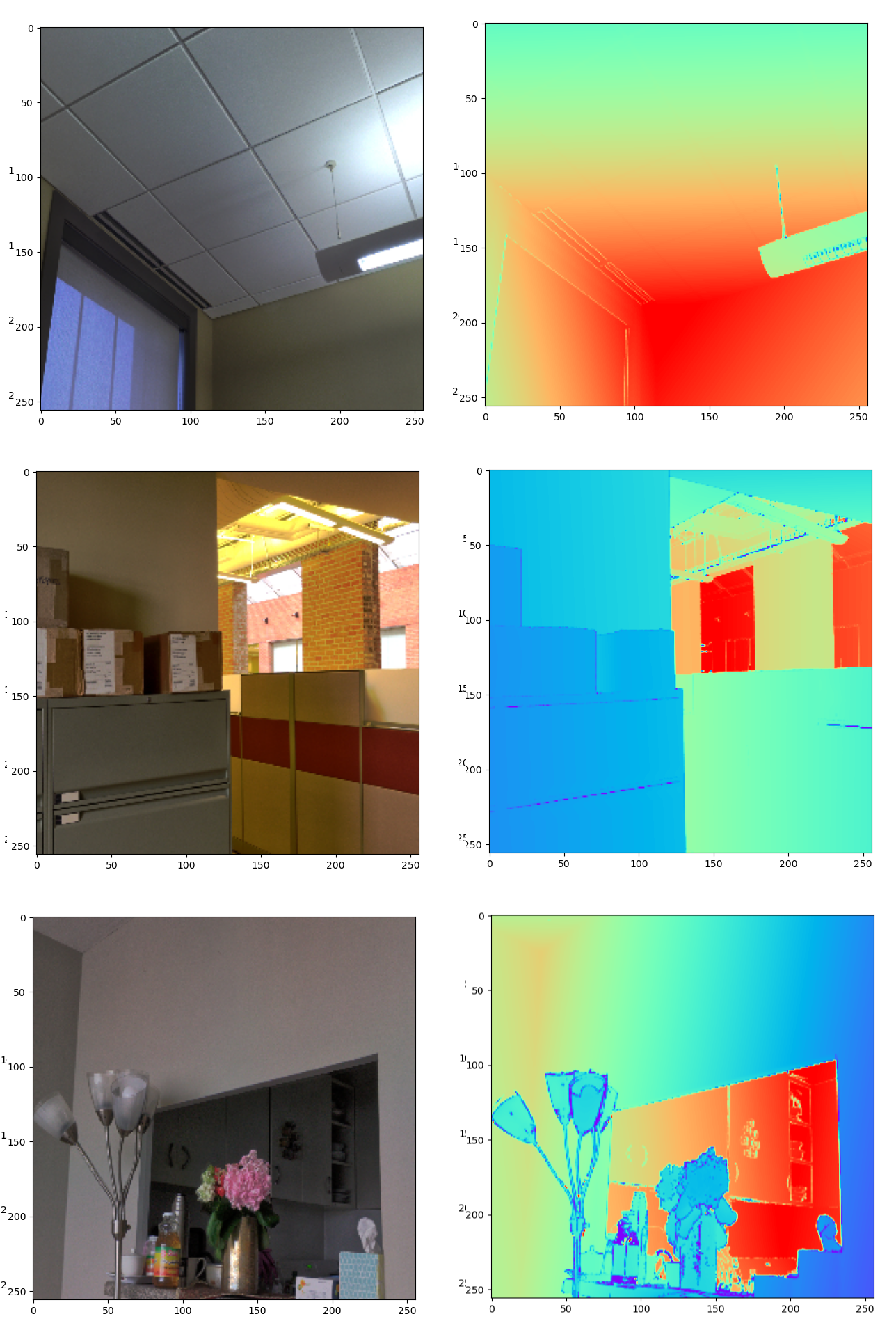}
	\caption{DIODE dataset.  Left:input and Right:Depth map}
	\label{fig:mono-dataset}
\end{figure}

In Fig \ref{fig:mono-dataset}, 3 samples are shown from DIODE dataset's test set.  The dataset was split between training and validation sets. Validation set was used to evaluate the model's performance on unseen data.

\begin{figure}[h]
	\centering
	\includegraphics[width=2.2in]{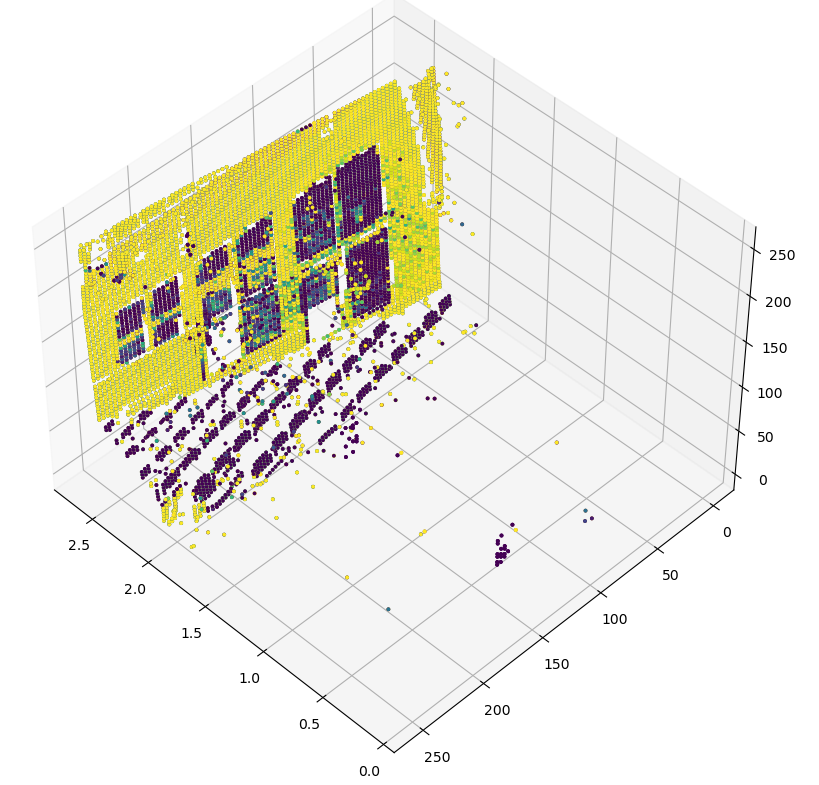}
	\caption{3D point cloud representation of input}
	\label{fig:mono-dataset-3d}
\end{figure}

Fig \ref{fig:mono-dataset-3d} shows a point cloud representation of an input image using its corresponding depth map.

\subsubsection{Stereo Dataset}
\begin{figure}[h]
	\centering
	\includegraphics[width=\linewidth]{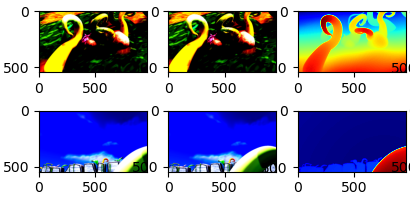}
	\caption{Monkaa Stereo dataset.  Left, Right input images and Left Disparity Map}
	\label{fig:stereo-dataset}
\end{figure}

KITTI \cite{Geiger2013IJRR}, NYU-V2 \cite{Silberman:ECCV12}, Scene Flow \cite{MIFDB16} are available datasets for stereo depth estimation. Each dataset is on the order of 100GB in size. We picked Monkaa, a subset of the Scene Flow dataset, as the target dataset for our study allowing us enough time to perform multiple ablation studies and model architecture evaluation. It contains ~9000 synthetic data samples and corresponding disparity maps. We split the data into train and test sets with a 90-10 split. The figure \ref{fig:stereo-dataset} shows reference example of a single data sample from the dataset.

\subsection{Baseline}
Our baseline is to reproduce the existing results as a first step. In the subsections below, we describe methods used to reproduce both Monocular and Stereo Vision based depth estimation techniques. We also show some ablation studies done to improve accuracy of predictions. 

\subsubsection{Monocular Depth}
We reproduced the chosen monocular depth estimation model and evaluated different types of loss functions.  Monocular depth estimation was built using Tensorflow and Keras framework \cite{tensorflow2015-whitepaper}.  The reference publication describes using structural similarity (SSIM), depth smoothness and L1 loss as three metrics. We performed training and hyper-parameter search for combination of these loss functions. Fig \ref{fig:mon_train_loss} and Fig \ref{fig:mon_val_loss} below show the convergence over time. We also varied batch sizes from 16, 32, 64. 

We used SSIM as an evaluation metric for measuring similarity between target and predicted data. 
Figure \ref{fig:mono-compare} shows results. 

\begin{figure}[h]
	\centering
	\includegraphics[width=\linewidth]{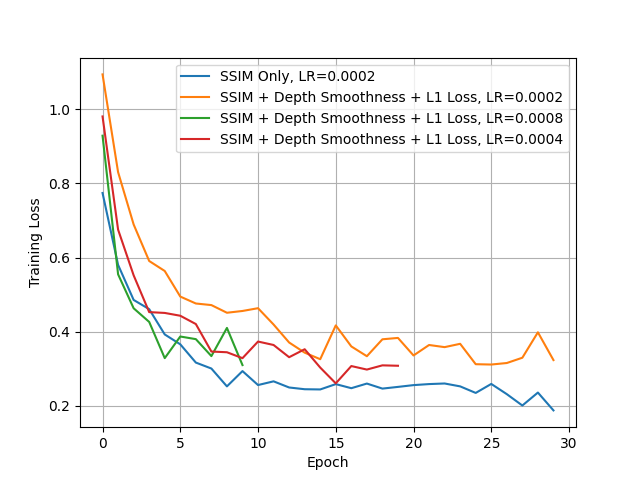}
	\caption{Monocular Depth - Train Loss}
	\label{fig:mon_train_loss}
\end{figure}

\begin{figure}[h]
	\centering
	\includegraphics[width=\linewidth]{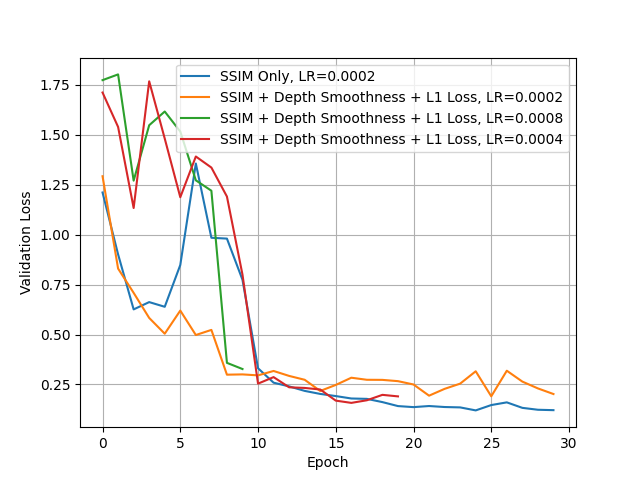}
	\caption{Monocular Depth - Validation Loss}
	\label{fig:mon_val_loss}
\end{figure}

\begin{figure}[h]
	\centering
	\includegraphics[width=2.5in]{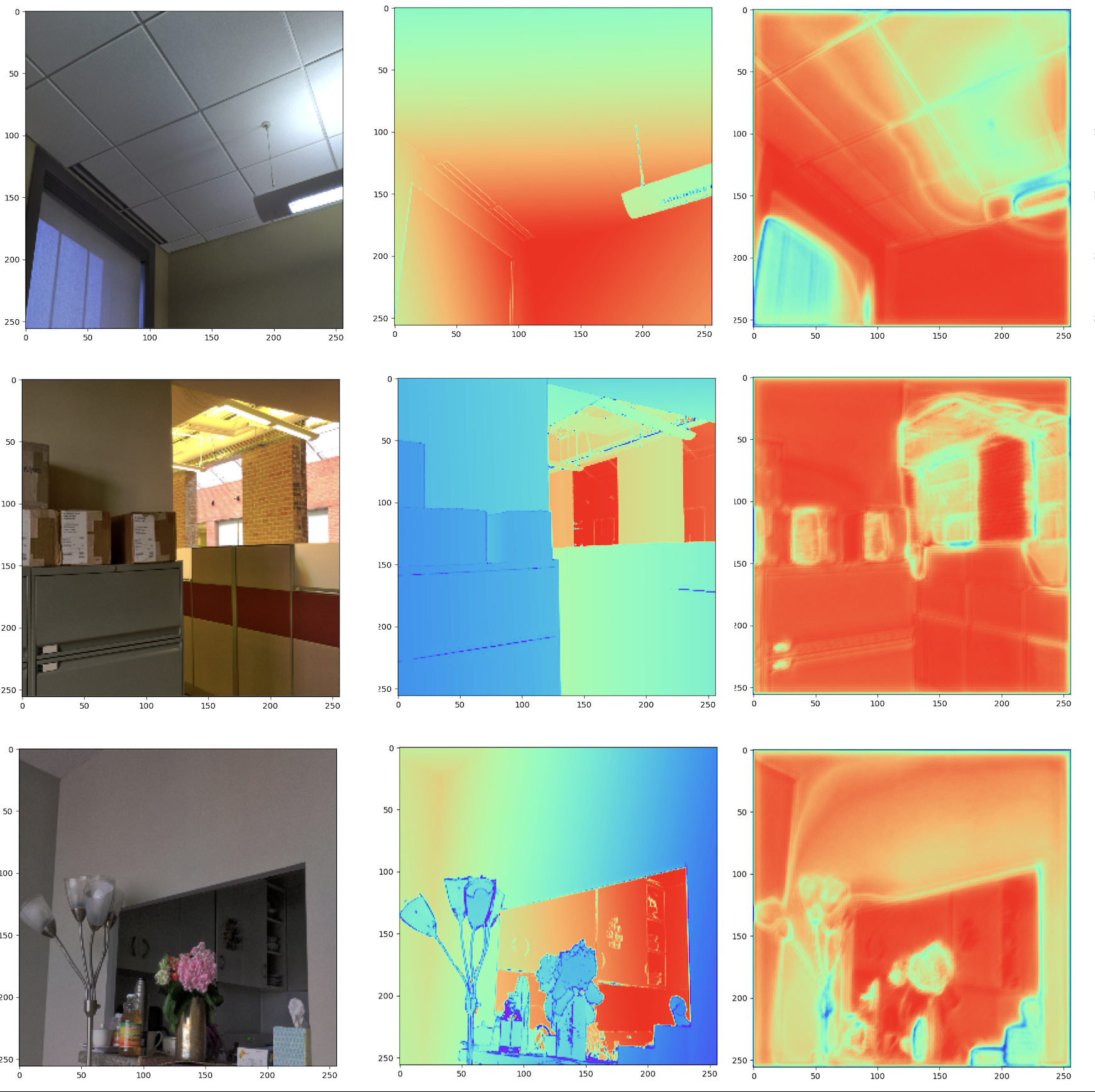}
	\caption{Monocular Depth - Input, Target, Predicted}
	\label{fig:mono-results-default}
\end{figure}

We observed that using SSIM as only loss metric resulted in better model convergence, as visualized by training and validation loss curves in figures \ref{fig:mon_train_loss} and \ref{fig:mon_val_loss}. However, visually, the results were not on part with using all three losses in optimization. We found this observation to align with our expectation. SSIM is generally lenient on errors in nearest neighbors, however, adding depth smoothness and L1 depth loss would provide a tighter correspondence between targets and predictions. 

\subsubsection{Stereo Depth - AnyNet}
\begin{figure}[h]
	\centering
	\includegraphics[width=\linewidth]{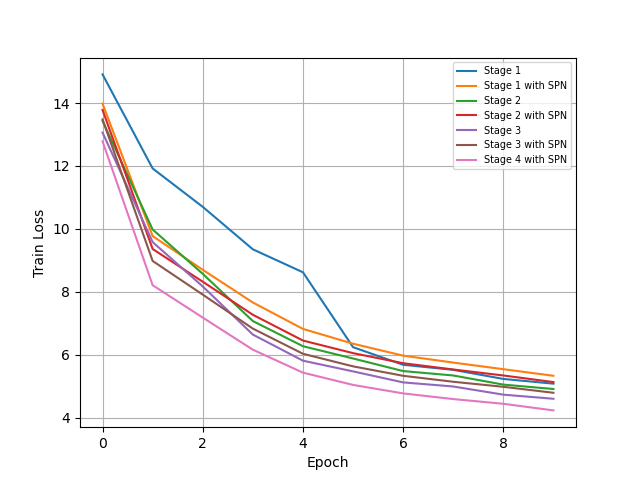}
	\caption{AnyNet - Train Loss}
	\label{fig:stereo_train_loss}
\end{figure}

\begin{figure}[h]
	\centering
	\includegraphics[width=\linewidth]{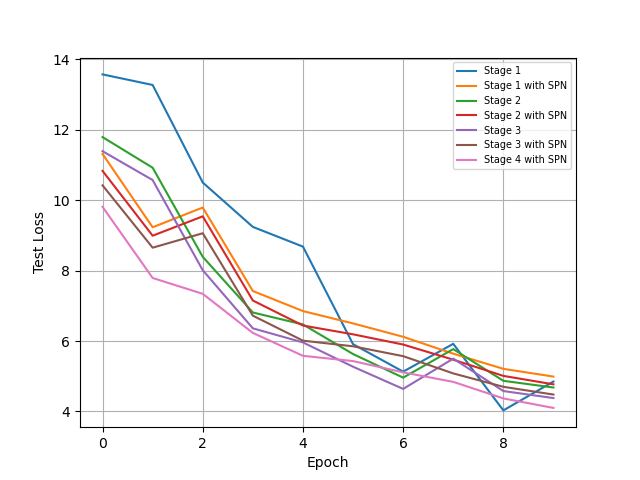}
	\caption{AnyNet - Test Loss}
	\label{fig:stereo_test_loss}
\end{figure}

We reproduced the AnyNet learning based stereo vision depth estimation model using Pytorch \cite{NEURIPS2019_9015}. Training process is computationally intensive and took ~4 hours on GTX1070 and ~30min on V100 GPUs. We were also able to use a larger dataset on this model compared to the Monocular depth model. The dataset was split 90\%-10\% for training and test sets randomly. Figure \ref{fig:stereo_train_loss} shows training convergence of 2 variations of the AnyNet model, with and without Spatial Propagation Network (SPNet). The publication \cite{anynet} describes that the model performs better with SPNet. Figures \ref{fig:stereo_test_loss} and \ref{fig:stereo_train_loss} shows losses at various stages, described in the model shown in figure \ref{fig:anynet_model}. The idea is that the user would be able to truncate the model at stage 1, 2, 3 with a slight loss in accuracy but a considerably smaller model size. 

The publication \cite{anynet} uses three pixel error for measuring disparity differences between target and predicted disparities. We have implemented the same metric to evaluate accuracy of the model.

\begin{figure}[h]
	\centering
	\includegraphics[width=\linewidth]{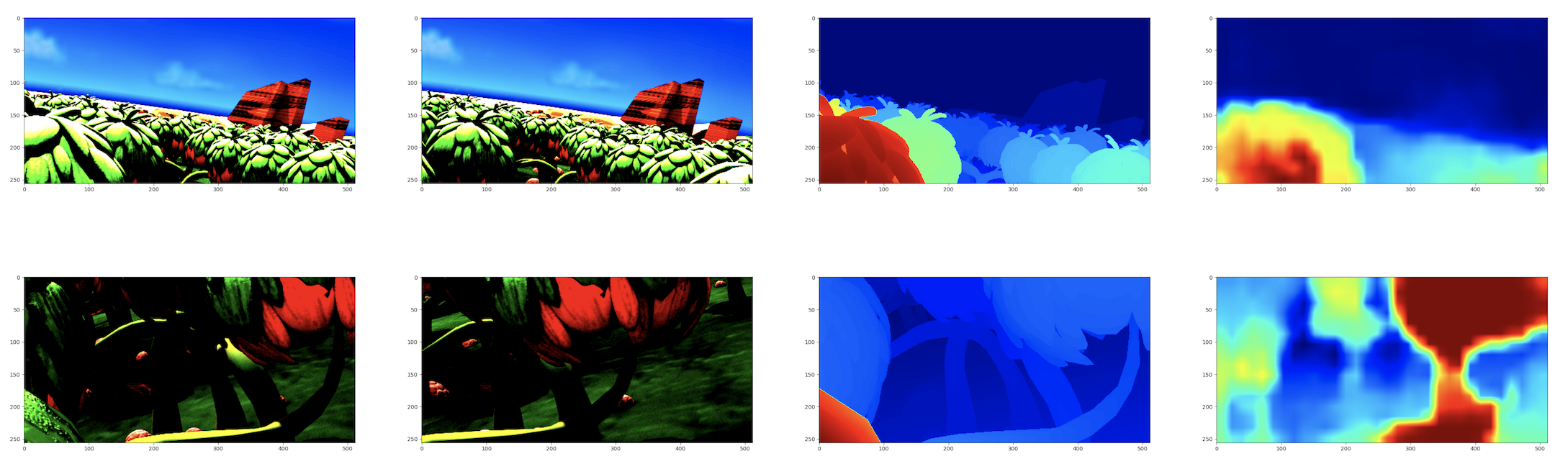}
	\caption{AnyNet - Left, Right inputs; Target \& Predicted disparity}
	\label{fig:stereo_results_default}
\end{figure}

Fig \ref{fig:stereo_results_default} shows predicted disparity map. Three pixel error accuracy is shown in the figure \ref{fig:stereo-box-plot} and table \ref{tab:stereo-models}. 

\subsection{Complexity Reduction}
Once we established our baseline with both the models, our subsequent goal was to evaluate computational complexity of the baseline models and investigate methods to reduce the model's computational complexity. Model size and number of parameters are our metrics to measure reduction in compute usage. On an embedded device mounted on a battery power autonomous system, reduction in model size can boost the efficiency of the whole system. Our evaluation metric remains the same, i.e. lower loss, lower SSIM for Monocular Depth model and lower three pixel error for Stereo Depth. 

The subsections below describe experiments and results on reducing computational complexity of the models. 

\subsubsection{Monocular Depth}
Monocular depth model has an autoencoder structure with 4 Down-scale blocks, 1 bottleneck block and 4 Up-scale blocks. Table \ref{tab:mono-structure} describes structure of the model (Conv2D is a 2-D convolution; BN is batch normalization; Act is activation function). 

\begin{table}
\begin{center}
\caption{Monocular Depth - Model structure }
\begin{tabular}{ |c|c|c|c| } 
\hline
 Block & Structure \\
 \hline
 Down-scale & 2x(Conv2D-BN-Act)+Pool \\
\hline
 Bottleneck & 2x(Conv2D-Act) \\
 \hline
 Up-scale & 2x(Conv2D-BN-Act) \\
\hline
\end{tabular}
\label{tab:mono-structure}
\end{center}
\end{table}

The filter sizes for the monocular model described in table \ref{tab:mono-structure} are in increased order 16-32-64-128-256. The higher order down-scale and upscale blocks are computationally expensive. Experiments were done in order to reduce the dependency on higher order filters and evaluate model accuracy, with reference to the baseline. 

The table below \ref{tab:mono-structure} describes three model variants, the first row  (Model-1) is the default monocular depth model. The second row (Model-2) is model variant by removing the higher order down scale and upscale blocks. This variation reduces the number of model parameters drastically, as shown in the table. In both Model-1 and Model-2, the activation function used is default LeakyRelU(0.2). Activation function was changed to Swish based on \cite{DBLP:journals/corr/abs-1710-05941} to investigate if it improves accuracy. This variant is shown in last row (Model-2 Swish) of the table \ref{tab:mono-models}. The last column of the table shows SSIM evaluation between target depth map and predicted depth map after training for 20 epochs and model convergence was achieved. 

The number of model parameters reduces from 1966467 to 489091, which is a 75.17\% reduction in model size!

\begin{table}
\begin{center}
\caption{Monocular Depth - Model variants, sizes \& accuracy }
\begin{tabular}{ |c|c|c|c| } 
\hline
 Model Structure & Activation & Parameters & Test SSIM\\
 \hline
 4-1-4 & LeakyReLU & 1966467 & 0.9895 \\
\hline
 3-1-3 & LeakyReLU & 489091 & 0.9903 \\
 \hline
 3-1-3 & Swish & 489091  & 0.9871 \\
\hline
\end{tabular}
\label{tab:mono-models}
\end{center}
\end{table}

For the given dataset, reducing the model structure from 4-1-4 to 3-1-3 reduced model parameters by more than two thirds and yet maintained relatively acceptable performance with respect to baseline. 

\begin{figure}[h]
	\centering
	\includegraphics[width=\linewidth]{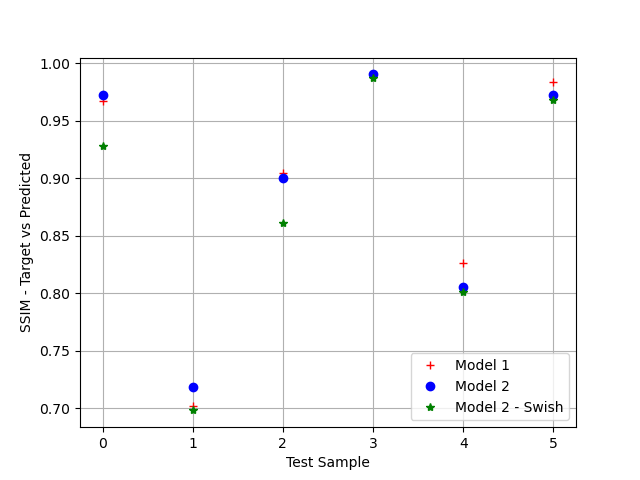}
	\caption{Monocular Depth - Models Comparison}
	\label{fig:mono-compare}
\end{figure}

\begin{figure}[h]
	\centering
	\includegraphics[width=\linewidth]{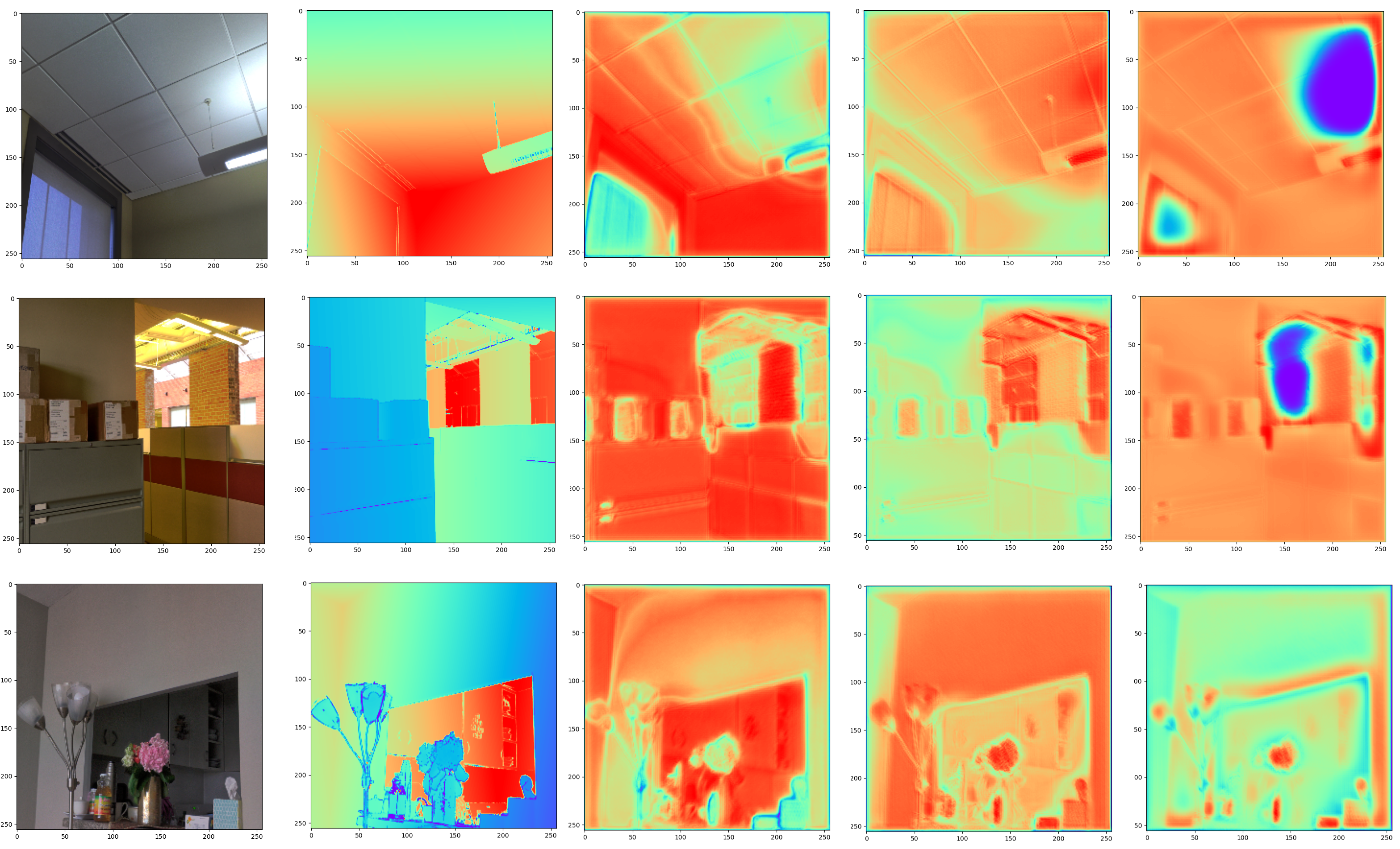}
	\caption{Monocular Depth - Input, Target, Model-1, Model-2, Model-2 Swish}
	\label{fig:mono-compare-vis}
\end{figure}

The figure \ref{fig:mono-compare} shows SSIM comparison of 6 random data samples from test set. The figure \ref{fig:mono-compare-vis} shows visual comparison of the depths maps obtained by 3 different models from \ref{tab:mono-models}. 

The figure \ref{fig:mono-compare-vis} shows visual data for the test samples 0, 2, 3, whose corresponding SSIM is shown in the figure \ref{fig:mono-compare}.

Based on this, we can conclude that the model complexity can be reduced without much impact to predicted depth map as shown by Model-1 vs Model-2. An interesting observation is that replacing LeakyReLU with another advanced activation function Swish did not improve the accuracy of the model. 

Another important observation can be made that although SSIM measured on the 3 models, as shown in \ref{tab:mono-structure} and \ref{fig:mono-compare} is close to 1 and very close to each other, the depth maps shown by figure \ref{fig:mono-compare-vis} show huge difference in qualitative perception of depth map. This can explain that SSIM alone cannot be a good loss function for training the Monocular Depth Estimation model. Thus, we can conclude that although the models converged acceptably during initial baseline ablation studies, predicted depth maps did not resemble target.

\subsection{Stereo Depth - AnyNet}
AnyNet model's structure is mainly comprised of 3 sections as shown in \ref{fig:anynet_model}. First section is a UNet for feature detection of left and right images separately. Second section is a 4D cost volume reduction. Third section is Spatial Propagation Network (SPN). SPN has larger number of channels in the filters. As UNet is crucial part of the model to detect disparity features, this is not primary focus to reduce complexity of the model. Furthermore, \cite{anynet} concretely suggests using larger models instead of using UNet if more accuracy is needed at the expense of larger model. Thus, we started by studying effect of SPN on the model accuracy. 

Similar to exercise done on previous model, our metric of reducing complexity of AnyNet is to reduce the number of model parameters, which in turn is a direct metric of reduction in compute and memory resources required. 

\begin{figure}[h]
	\centering
	\includegraphics[width=\linewidth]{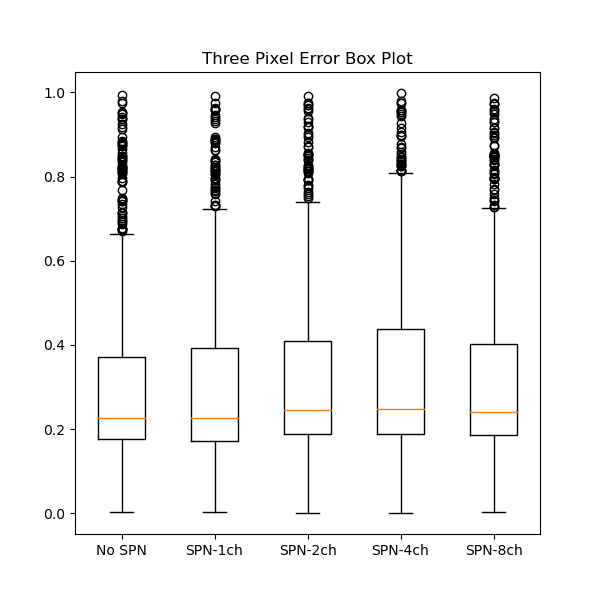}
	\caption{AnyNet - Three Pixel Error on test set for 5 model variants }
	\label{fig:stereo-box-plot}
\end{figure}

We compared 5 model variants of AnyNet in Table \ref{tab:stereo-models}. Each model's total number of parameters and average three pixel error across entire test are shown in the table. The lesser the three pixel error, the higher the accuracy of the model. Fig \ref{fig:stereo-box-plot} shows distribution of three pixel error across all the samples in test set (~800 samples). We do observe some outliers but it is promising to see that all variants of the model have very low error.  

\begin{table}[h]
\begin{center}
\caption{AnyNet - Model variants, sizes \& accuracy} 
\label{tab:stereo-models}
\begin{tabular}{ |c|c|c|c| } 
\hline
 Model Structure & Parameters & Three Pixel Error\\
 \hline
 No SPN  & 34629 & 0.2994 \\
 \hline
 SPN 1 channel  & 34827 & 0.3048 \\
 \hline
 SPN 2 channels & 35277  & 0.3193 \\
 \hline
 SPN 4 channels & 36933  & 0.3264 \\
 \hline
 SPN 8 channels & 43269  & 0.3178 \\ 
\hline
\end{tabular}
\end{center}
\end{table}

\begin{figure}[h]
	\centering
	\includegraphics[width=\linewidth]{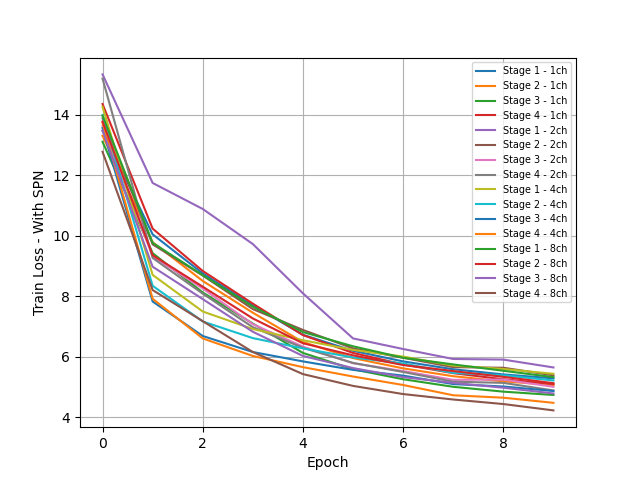}
	\caption{AnyNet - Train Loss for 5 model variants }
	\label{fig:stereo-compare-train}
\end{figure}

\begin{figure}[h]
	\centering
	\includegraphics[width=\linewidth]{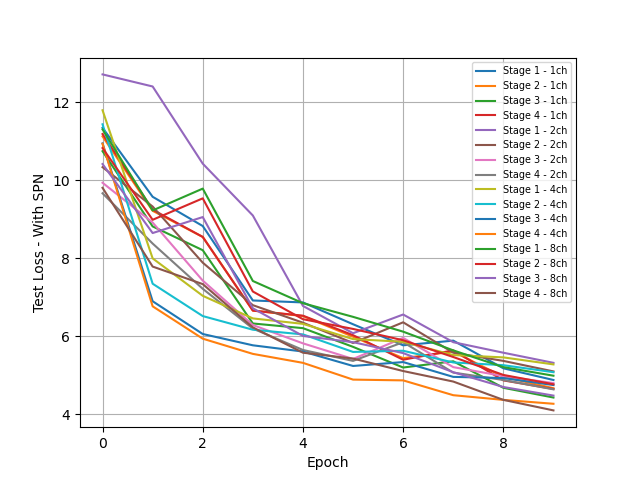}
	\caption{AnyNet - Test Loss for 5 model variants }
	\label{fig:stereo-compare-test}
\end{figure}

The figures \ref{fig:stereo-compare-train} and \ref{fig:stereo-compare-test} show training and test loss convergence for these 5 model variants. 

\begin{figure}[h]
	\centering
	\includegraphics[width=\linewidth]{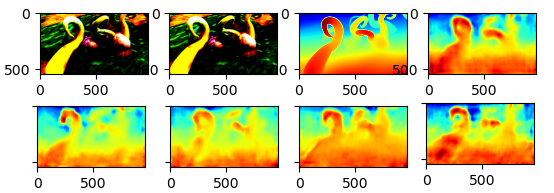}
	\caption{AnyNet - Test Sample 0 - Top Row (Left to Right): Left input, Right input, Target Disparity, Disparity with No SPN Model; Bottom Row (Left to Right): Disparity with 1 channel SPN, Disparity with 2 channel SPN, Disparity with 4 channel SPN, Disparity with 8 channel SPN }
	\label{fig:stereo-test0-no01248}
\end{figure}

\begin{figure}[h]
	\centering
	\includegraphics[width=\linewidth]{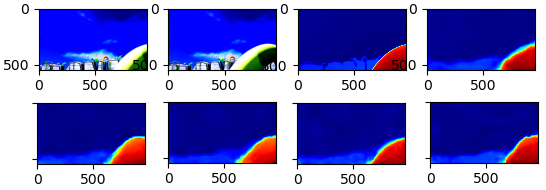}
	\caption{AnyNet - Test Sample 1 - Top Row (Left to Right): Left input, Right input, Target Disparity, Disparity with No SPN Model; Bottom Row (Left to Right): Disparity with 1 channel SPN, Disparity with 2 channel SPN, Disparity with 4 channel SPN, Disparity with 8 channel SPN }
	\label{fig:stereo-test1-no01248}
\end{figure}

Figures \ref{fig:stereo-test0-no01248} and \ref{fig:stereo-test1-no01248} showcase visually the disparity maps obtained by training the 5 variants of AnyNet models. As observed, the difference is relatively low on model with 1 channel SPN vs the model with 8 channel SPN, with ~20\% lesser model size as shown by the distribution of the 25th and 75th quartiles.

From the before mentioned experiments we can conclude that addition of SPN  improves accuracy slightly compared to without SPN. As shown in \ref{tab:stereo-models} we can choose 1 channel SPN based AnyNet instead of 8 channel SPN with ~20\% lesser model size. 

%-------------------------------------------------------------------------
%------------------------------------------------------------------------
\section{Conclusion}
As an initial step, we reproduced learning based methods to estimate depth using both, Monocular and Stereo Vision techniques. We performed ablation studies to tune hyper parameters such as learning rate in order to improve accuracy. We investigated on using various loss functions on monocular depth estimation. After establishing baselines in both methods, we experimented with modifying model architecture by changing the neural network layer sizes. We also changed activation functions to improve accuracy after reducing network size. 

For stereo depth estimation, we started from the novel state-of-the-art approach of \cite{anynet} and studied advantages of reducing the spatial propagation network in reducing model complexity. 

In both the methods, we established that reducing model size is a viable option to reduce computational complexity, that consequently reduces energy consumption for deploying these techniques in highly constrained navigation systems. To further improve quality of experiments, future work can use same dataset on both monocular and stereo depth techniques for a quantitative and qualitative comparison of both models. However, stereo technique may always outperform monocular depth technique in accuracy. 

After reducing the model sizes, we also tried using multi-precision casting in both Pytorch and Tensorflow but were unsuccessful in converting the model from a 64b/32b to a bfloat16 or float16 model. This could be a reasonable future work that can be undertaken as reducing the data type from 32b to 16b further reduces compute and memory needs by 50\%. For the most part, this would only need fine-tuning the model. Similarly, the models can be further reduced computationally by pruning and quantizing but this would hurt the accuracy furthermore. Thus, quantizing and pruning may need additional investigative work (FastDepth \cite{wofk2019fastdepth} research used pruning in their method).

{\small
\bibliographystyle{ieee}
\bibliography{ref}
}

\end{document}